\documentclass[sigconf,final]{acmart}

\acmConference[EASE 2026]{The 30th International Conference on Evaluation and Assessment in Software Engineering}{9–12 June, 2026}{Glasgow, Scotland, United Kingdom}

\AtBeginDocument{%
  }
    
\usepackage{hyperref}
\usepackage{multirow}
\usepackage{enumitem}
\usepackage{listings}
\usepackage{color}
\usepackage{tcolorbox}
\usepackage{amsmath}
\usepackage{algorithmic}
\usepackage{graphicx}
\usepackage{textcomp}
\usepackage[table,xcdraw]{xcolor}
\usepackage{xspace}
\definecolor{blue}{HTML}{2f4b7c}

\definecolor{dkgreen}{rgb}{0,0.6,0}
\definecolor{gray}{rgb}{0.5,0.5,0.5}
\definecolor{mauve}{rgb}{0.58,0,0.82}
\definecolor{lightred}{RGB}{237, 67, 55}
\definecolor{lightviolet}{RGB}{215, 75, 118}
\newboolean{showcomments}
\setboolean{showcomments}{true}
\ifdefined\notodocomments
  \setboolean{showcomments}{false}
\fi
\ifthenelse{\boolean{showcomments}}
 { \newcommand{\mynote}[2]{
      \fbox{\bfseries\sffamily\scriptsize#1}
        {\small$\blacktriangleright$\textsf{\emph{#2}}$\blacktriangleleft$}}}
        { \newcommand{\mynote}[2]{}}

\begin{document}

\title{LeGo-Code: Can Modular Curriculum Learning Advance Complex Code Generation? Insights from Text-to-SQL}
\author{Salmane Chafik}
\affiliation{%
 \institution{Mohammed VI Polytechnic University}
 \country{Morocco}}
\author{Saad Ezzini}
\affiliation{%
 \institution{King Fahd University of Petroleum and Minerals }
 \country{Saudi Arabia}}
 \author{Ismail Berrada}
\affiliation{%
 \institution{Mohammed VI Polytechnic University}
 \country{Morocco}}

\begin{abstract}

Recently, code-oriented large language models (LLMs) have demonstrated strong capabilities in translating natural language into executable code. Text-to-SQL is a significant application of this ability, enabling non-technical users to interact with relational databases using natural language. However, state-of-the-art models continue to struggle with highly complex logic, particularly deeply nested statements involving multiple joins and conditions, as well as with real-world database schemas that are noisy or poorly structured. In this paper, we investigate whether curriculum learning can improve the performance of code-based LLMs on Text-to-SQL tasks. Employing benchmarks including Spider and BIRD, we fine-tune models under different curriculum strategies. Our experiments show that naïve curriculum, simply ordering training samples by complexity in a single epoch, fails to surpass standard fine-tuning due to catastrophic forgetting. To overcome this, we propose a Modular Adapter Composition (MAC) strategy. By sequentially training tier-specific adapters on incremental complexity levels (Easy to Extra-Hard), we create a scaffolded learning environment that improves performance on complex queries. Our approach not only produces measurable performance gains on the Spider and BIRD benchmarks but also provides a flexible, "Lego-like" architecture, allowing models to be composed and deployed based on specific schema difficulty requirements. These findings demonstrate that structured, modular learning is a superior alternative to monolithic fine-tuning for mastering the syntax and logic of complex code generation.

\end{abstract}



\maketitle

\section{Introduction}

During the last few years, the software engineering domain has benefited greatly from the introduction of large language models (LLMs) for code. Software developers now rely heavily on these tools in their daily activities, including generating ideas, designing application architectures, writing code snippets, testing code, maintaining projects, debugging and securing software. Today, tools such as Claude Code \cite{claudecode} and Cursor \cite{cursor} can generate and manage even large repositories on GitHub. Since the introduction of the Transformer architecture \cite{vaswani2017attention}, software engineering tasks leaderboards have been consistently dominated by code-focused LLMs.

Structured Query Language (SQL) is one domain that has been highly influenced by these advances. Today, non-technical users can interact with relational databases using natural language rather than writing formal SQL queries. Users can ask questions in everyday language, often in multiple languages, not limited to English, without prior knowledge of SQL syntax, schema design, or database constraints. In response, AI systems translate these requests into executable SQL statements that retrieve the desired information. This domain, commonly referred to as Text-to-SQL, enables more accessible data exploration and democratizes database usage across organizations. Instead of relying solely on database specialists or analysts, business users, researchers, and decision-makers can directly query data sources through conversational interfaces. Modern systems can also incorporate schema awareness, contextual understanding, and ambiguity resolution to produce accurate and efficient queries \cite{song2024enhancing}.

With the introduction of multiple datasets and benchmarks, such as WikiSQL \cite{wikisql}, SPIDER \cite{spider}, and BIRD \cite{bird}, the Text-to-SQL field has witnessed remarkable progress over time. Early approaches relied primarily on rule-based systems, template matching, or sketch-based methods \cite{sketch_0}. These techniques relied on predefined query templates or structures with placeholders that were automatically filled based on the user’s input. However, their reliance on rigid templates made them inflexible and poorly suited to handle linguistic variability, ambiguous phrasing, or complex query requirements. To address these limitations, subsequent research explored graph-based approaches \cite{graph_3} and the use of intermediate representations and languages\cite{inter_0}. These methods aimed to better capture the structural relationships between natural language questions and database schemas, improving robustness and generalization beyond fixed templates. More recently, modern approaches have leveraged Transformer-based architectures\cite{qwencoder, starcoder}, which excel at modeling long-range dependencies and deep semantic relationships in both natural language and code. Such models can generalize across domains, handle diverse linguistic expressions, and adapt to previously unseen database schemas. Consequently, Transformer-based methods have achieved state-of-the-art performance on benchmark datasets and enabled the practical deployment of natural-language interfaces to databases in real-world applications.

However, even with these advances, state-of-the-art models still struggle with highly complex SQL queries \cite{bird}. This is particularly evident for deeply nested queries involving multiple joins, aggregations, and conditional clauses. The challenge becomes even greater when models must operate on very large databases containing noisy or inconsistent data, poorly normalized schemas, or non-descriptive attribute and key names. These real-world conditions introduce ambiguity and reasoning difficulties that current systems have yet to fully overcome.

Motivated by these limitations, this work investigates whether curriculum learning can enhance the ability of code-oriented LLMs to handle complex Text-to-SQL cases. We focus particularly on queries that involve sophisticated logical reasoning, such as deeply nested subqueries, multiple table joins, and intricate filtering conditions, as well as scenarios involving large, noisy, or poorly structured real-world schemas, exactly as the samples present in the BIRD benchmark.

Our primary contributions are the following: 

\begin{itemize}

\item A comparative study of direct sorting, single-stage fine-tuning, and multi-stage curriculum learning approaches.

\item \textbf{LeGo-Code}, A modular training approach using multiple adapters specialized for different query complexities.

\item An empirical evaluation defining the contexts in which curriculum learning yields improvements for Text-to-SQL tasks, alongside an analysis of its limitations.

\end{itemize}

The remainder of this paper is organized as follows: Section \ref{Sec:Related} reviews related work relevant to text-to-SQL. Section \ref{Sec:Approach} details the different curriculum learning strategies we employed for text-to-SQL. Section \ref{Sec:Eval} presents the experimental setup and comparative analyses against direct finetuning. Section \ref{Sec:Discussion} discusses the key findings, and finally, Section \ref{Sec:Conclusion} concludes the paper and outlines potential directions for future research.
\section{Related Work}\label{Sec:Related}

Translating natural language questions into their correct executable SQL queries, while respecting the database schema and capturing the true intent of the request, requires a deep understanding of natural language to resolve ambiguities, as well as strong knowledge of code, SQL syntax, data types, constraints, and database structures.

Prior work has explored sketch-based slot-filling methods \cite{sketch_0, sketch_1}. These approaches convert SQL generation into multiple classification tasks by utilizing predefined SQL query templates and predicting their components sequentially. While decomposing the text-to-SQL task into subtasks facilitates the generation of syntactically correct SQL queries, it also introduces several limitations, particularly reduced flexibility in handling diverse SQL query structures due to the rigid nature of the predefined sketches. Others focused on Graph based methods \cite{graph_0, graph_1}, employing either Graph Neural Networks (GNNs) or modified transformer architectures. These methods represent the database schema as a graph, capturing relationships between tables, their columns, and inter-table dependencies. By incorporating structural information through graph representations, researchers achieved modest performance improvements, though these gains often fell short of practical requirements.

In the same context, other works have explored the use of intermediate languages \cite{inter_0, inter_1}, either pre-existing ones or new ones, specifically designed for the text-to-SQL task. These intermediate representations are typically closer to natural language, providing a bridge between the input question and the target SQL query. By translating natural language into an intermediate form first, and then converting it into SQL, these methods aim to simplify the generation process and improve semantic alignment. The effectiveness of these methods depends heavily on the expressiveness of the intermediate language and the accuracy of the mapping from intermediate forms to executable SQL. 

With the introduction of several large-scale datasets, the field of text-to-SQL has progressed significantly. Notable benchmarks include WikiSQL \cite{wikisql}, which provides NLQ–SQL pairs over single tables; Spider \cite{spider}, the first handcrafted dataset featuring substantially more complex queries than WikiSQL, including multi-table joins and nested SQL statements; and BIRD \cite{bird}, a challenging benchmark that introduces very large database schemas, dirty values, and real-world scenarios. BIRD contains highly complex queries, often involving up to eight joins, deeply nested structures, sophisticated aggregation functions, and statistical computations, making it the most demanding benchmark to date. Together with numerous multilingual and synthetic datasets \cite{data_1, data_4, data_2, data_5, data_3, data_6}, these resources have provided the foundation for code-oriented LLMs \cite{starcoder, qwencoder, codellama} to achieve state-of-the-art performance and dominate text-to-SQL leaderboards.

Recent research \cite{prompt_1, prompt_2} has demonstrated that prompt engineering can significantly improve the performance of LLMs on the text-to-SQL task by better guiding the model’s reasoning process. These approaches leverage in-context learning, where carefully selected demonstrations are provided at inference time, as well as chain-of-thought (CoT) prompting strategies that encourage step-by-step reasoning before generating the final SQL query. Similarly, other researchers have sought to enhance LLM performance without modifying model architectures, particularly through curriculum learning. This paradigm trains models progressively, starting with simpler text-to-SQL examples and gradually increasing difficulty to medium and highly complex queries. 

Recent work \cite{cl_1} explored curriculum learning for text-to-SQL using a synthetic dataset generated by both closed-source and open-source LLMs. Their experiments reported performance improvements when adopting a curriculum-based training strategy. However, the reliance on synthetically generated data, produced through prompting heuristics and LLM outputs, may introduce distributional biases and may not fully capture the diversity and realism of user-generated queries. Furthermore, their study employed a single-stage curriculum learning strategy, leaving open the question of how different curriculum designs might influence model generalization and robustness.

Overall, prior work has improved text-to-SQL through structured decoding, graph modeling, intermediate languages, and prompting strategies. However, the relative effectiveness of different training paradigms for handling varying query complexities remains underexplored. In particular, there is limited systematic comparison between direct fine-tuning, and curriculum learning strategies, and little investigation into modular training methods tailored to query difficulty. To address these gaps, we conduct a comparative study of these training approaches, introduce LeGo-Code, a modular adapter-based framework specialized for different query complexities, and provide an empirical analysis of when curriculum learning benefits text-to-SQL.

\section{Approach} \label{Sec:Approach}

This section provides a detailed description of the datasets used, the sorting algorithm applied to organize samples from easy to complex, and the training strategies employed.

\subsection{Datasets}

In this work, we employ two cross-domain text-to-SQL benchmarks, SPIDER and BIRD, which differ totally in complexity, domain coverage, database scale, and real-world applicability. Together, they provide a comprehensive evaluation setting across varying levels of difficulty and practical use cases.

\textbf{SPIDER}: A large-scale cross-domain text-to-SQL dataset containing 10,181 NLQ-SQL pairs corresponding to 5,693 unique SQL queries. The dataset spans 138 distinct domains and covers more than 200 databases, each consisting of multiple interrelated tables with complex schemas and numerous foreign-key relationships. SPIDER is the first large-scale hand-crafted benchmark for cross-domain semantic parsing, created by 11 computer science students to ensure high annotation quality and diversity. Its design requires models to generalize to unseen databases at test time, making it a challenging and widely adopted benchmark for both training and evaluation. The name “SPIDER” reflects its broad coverage across many domains, analogous to a spider traversing multiple interconnected nests (databases).

\textbf{BIRD}: is a more recent text-to-SQL benchmark designed to reflect higher complexity and closer alignment with real-world database scenarios. It contains 12,751 unique NLQ–SQL pairs over 95 large-scale databases with a total size of approximately 33.4 GB. The dataset spans more than 37 professional domains, including blockchain, sports (e.g., hockey), healthcare, education, and finance. Unlike earlier benchmarks, BIRD emphasizes real-world characteristics: databases and queries are derived from practical applications and therefore include noisy inputs, incomplete or inconsistent values, abbreviations, domain-specific terminology, and complex schema structures. This makes BIRD particularly suitable for evaluating robustness, scalability, and practical deployment readiness of text-to-SQL systems.these use cases.  

\begin{table}[]
\footnotesize 
\begin{tabular}{|l|c|c|c|c|c|c|}
\hline
\rowcolor[HTML]{FFFE65} 
\textbf{Dataset} & \textbf{Subset} & \textbf{N° Ex} & \textbf{N° DB} & \textbf{N° Tab/DB} & \textbf{N° Col/TB} & \textbf{N° FK/DB} \\ \hline
\cellcolor[HTML]{FFFC9E}                                  & Train & 7 000 & 140 & 5.26 & 5.22  & 4.9  \\ \cline{2-7} 
\multirow{-2}{*}{\cellcolor[HTML]{FFFC9E}\textbf{SPIDER}} & Dev   & 1 034 & 20  & 4.05 & 5.44  & 3.2  \\ \hline
\cellcolor[HTML]{FFFC9E}                                  & Train & 6 604 & 69  & 5.68 & 7.15  & 8.16 \\ \cline{2-7} 
\multirow{-2}{*}{\cellcolor[HTML]{FFFC9E}\textbf{BIRD}}   & Dev   & 1 534 & 11  & 6.81 & 10.64 & 9.27 \\ \hline
\end{tabular}
\caption{Schema Complexity Comparison of SPIDER and BIRD Datasets.}
\label{tab:structural_statistics}
\vspace{-30pt}
\end{table}

Table \ref{tab:structural_statistics} presents schema-level statistics for the training and development subsets of SPIDER and BIRD, including dataset size and average structural properties of the underlying databases. Metrics such as tables per database, columns per table, and foreign keys per database provide insight into the relational complexity faced by text-to-SQL models. These statistics indicate that BIRD database schemas are generally more complex than those in SPIDER. For example, the BIRD development set contains databases with nine relationships on average, approximately seven tables, each with around eleven columns. In contrast, the SPIDER development set contains databases with only about three relationships, four tables, each with roughly five columns. This highlights the higher structural complexity and real-world difficulty of the BIRD benchmark compared to SPIDER. Additional complexity statistics are presented in the following subsection.

\subsection{Curriculum Learning}

Curriculum Learning (CL) is a method used in Machine Learning (ML) where training data is presented to the model in a specific order, typically from easy to complex examples. This approach is inspired by how humans learn, starting with simpler concepts and gradually progressing to more challenging ones. In standard machine learning, models usually receive training data in a random order. With curriculum learning, the model begins with simpler examples and progressively moves to harder ones, which can improve learning efficiency and lead to more accurate predictions. 

One major challenge in Curriculum Learning is defining what makes a training example “easy” or “hard.” However, in text-to-SQL tasks, it is relatively straightforward to assess the complexity of SQL queries based on their components. For example, queries that join multiple tables are more complex than those that access only a single table. To quantify this, we define a complexity score function $C(Q)$ that takes a query $Q$ as input and outputs a real-valued score computed as a weighted sum over its key components. We extended the component weights from the work of \cite{cl_1} to include condition operators, aggregate functions, and, most importantly, the database size. Queries operating over larger database schemas are significantly more complex than those over smaller, simpler schemas. The detailed weights assigned to SQL keywords and functions are summarized in Table \ref{tab:sql_weights}. Nested SQL queries are detected using an SQL parser named SQLGlot\footnote{\url{https://github.com/tobymao/sqlglot}}, and the database size score $DB\_score$ is computed as $( db\_size * 2 ) / max\_size$ , where $db\_size$  is the number of characters in the database and $max\_size$  is the maximum database size in the dataset, resulting in a score between 0 and 2. The final complexity score is defined as follows:

\vspace{0.3cm}
$C(Q) = \sum_{k \in Q} n_k * w_k + DB\_score + Nested\_score(Q)$
\vspace{0.3cm}

\noindent Where $n_k$, $w_k$ are, respectively, the number of occurrences and the weight of the SQL keyword or function $k$, and $Nested\_score(Q) = 2$ if the query contains a nested SQL subquery (0 otherwise).

\begin{table}[]
\small
\begin{tabular}{|
>{\columncolor[HTML]{FFFC9E}}l |c|}
\hline
\cellcolor[HTML]{FFFE65}\textbf{Keyword / Function} & \cellcolor[HTML]{FFFE65}\textbf{Weight} \\ \hline
WHERE                  & \textbf{0.5}  \\ \hline
AND/OR/NOR             & \textbf{0.1}  \\ \hline
MAX/MIN/AVG/SUM        & \textbf{0.15} \\ \hline
COUNT/CAST/DISTINCT    & \textbf{0.15} \\ \hline
JOIN                   & \textbf{1.5}  \\ \hline
GROUP BY/HAVING        & \textbf{0.75} \\ \hline
ORDER BY/LIMIT         & \textbf{0.5}  \\ \hline
UNION/INTERSECT/EXCEPT & \textbf{2}    \\ \hline
Nested SQL Queries     & \textbf{2}    \\ \hline
Database Size                                       & \textbf{(db\_size*2)/max\_size}         \\ \hline
\end{tabular}
\caption{Weights assigned to SQL keywords and functions}
\label{tab:sql_weights}
\vspace{-25pt}
\end{table}

Following this, we merge the training sets of SPIDER and BIRD and sort the combined dataset using our scoring function, from easy to extra-hard samples. We denote the resulting dataset as \textbf{\texttt{SB-CL}}. For the subsequent training strategies, we require different subsets of this dataset; therefore, we partition \textbf{\texttt{SB-CL}} into four complexity sets of equal size (EASY, MEDIUM, HARD, and EXTRA). Summary statistics for these sets are presented in Table \ref{tab:complexity-levels}. These statistics highlight the complexity gap between SPIDER and BIRD, most SPIDER samples are concentrated in the EASY and MEDIUM sets, while BIRD samples are predominantly found in the HARD and EXTRA sets.

\begin{table}[]
\small
\begin{tabular}{|l|c|c|c|c|c|c|}
\hline
\rowcolor[HTML]{FFFE65} 
\textbf{Dataset} & \textbf{Subset} & \textbf{Easy} & \textbf{Medium} & \textbf{Hard} & \textbf{Extra} & \textbf{Avg\_Score} \\ \hline
\cellcolor[HTML]{FFFC9E}                         & Train & 2 561 & 1 957 & 1 113 & 1 369 & 1.982 \\ \cline{2-7} 
\multirow{-2}{*}{\cellcolor[HTML]{FFFC9E}SPIDER} & Dev   & 411   & 263   & 163   & 197   & 1.930 \\ \hline
\cellcolor[HTML]{FFFC9E}                         & Train & 840   & 1 441 & 2 288 & 2 032 & 2.780 \\ \cline{2-7} 
\multirow{-2}{*}{\cellcolor[HTML]{FFFC9E}BIRD}   & Dev   & 188   & 269   & 709   & 368   & 2.594 \\ \hline
\end{tabular}
\caption{Complexity Levels Comparison of SPIDER and BIRD.}
\label{tab:complexity-levels}
\vspace{-25pt}
\end{table}

\subsection{Training Strategies}

Once the dataset is prepared, merged, and sorted, we evaluate several training strategies to study the effect of sample complexity and curriculum design on model performance.

\subsubsection{LoRA Finetuning}

In this approach, we perform standard fine-tuning without considering the complexity order of the samples. The dataset is shuffled, and the model is trained for multiple epochs over the entire dataset. To improve efficiency, we adopt Parameter-Efficient Transfer Learning (PEFT) \cite{houlsby2019parameter}, specifically LoRA (Low-Rank Adaptation). In LoRA-based fine-tuning, the original model parameters are frozen, and only a small set of trainable adapter parameters is introduced. This significantly reduces computational cost and memory requirements compared to full fine-tuning, which is typically expensive for large language models.

This setup serves as the baseline for comparison with the curriculum-based methods.

\subsubsection{Single Stage Curriculum finetuning}

In this method, we apply a single-stage curriculum learning strategy. The model is fine-tuned on SB-CL, our pre-ordered dataset, where samples are arranged from easy to complex. Unlike the baseline, the dataset is not shuffled. Samples are presented sequentially to ensure that the model encounters simpler examples first and gradually progresses to more difficult ones. This ordering is intended to stabilize training and facilitate incremental knowledge acquisition. As in the baseline, we employ PEFT with LoRA adapters, keeping the base model frozen and training only the adapter parameters.

\subsubsection{Multi-Adapter based Curriculum finetuning}

This approach implements a multi-stage curriculum using multiple adapters. The SB-CL dataset is divided into four complexity groups: EASY, MEDIUM, HARD, and EXTRA. Training proceeds sequentially across stages:

\begin{enumerate}
    \item Train Adapter $1$ on the EASY subset while freezing the base model parameters.
    \item Freeze the base model and Adapter $1$, then train Adapter $2$ on the MEDIUM subset.
    \item Freeze the base model and the first two adapters, then train Adapter $3$ on the HARD subset.
    \item Freeze the base model and the first three adapters, then train Adapter $4$ on the EXTRA subset.
\end{enumerate}

\begin{figure*}[htbp]
    \centering
    \includegraphics[width=1\textwidth]{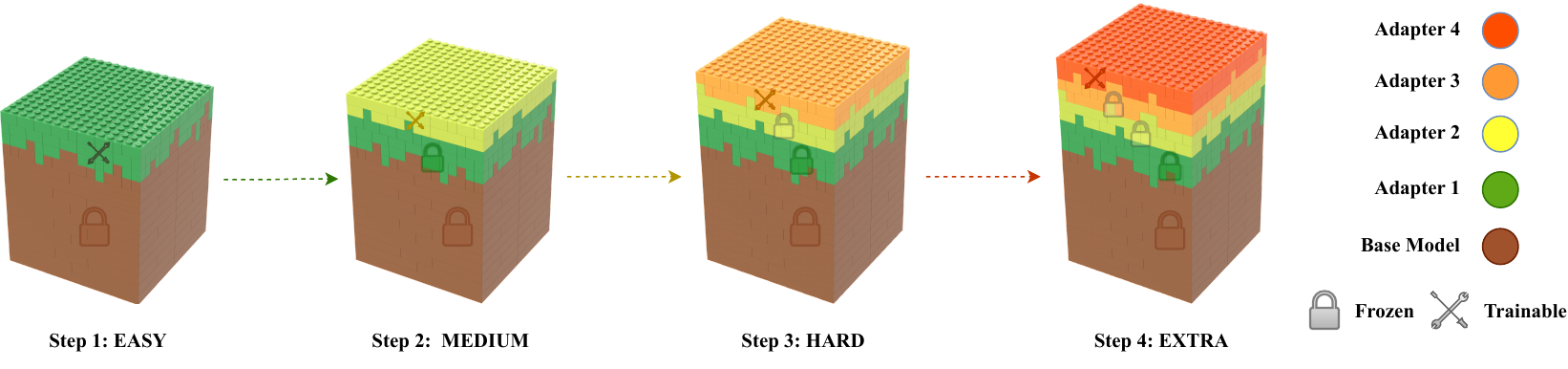}
    \caption{Multi-adapter Curriculum finetuning}
    \label{fig:lego-code}
\end{figure*}

At the end of this process, the final system consists of the base model augmented with four adapters, each specialized for a specific complexity level. Because each adapter is trained on top of the previously learned representations, later stages benefit from knowledge acquired in earlier ones. This design offers substantial flexibility at inference time. Different adapters can be enabled, disabled, or combined depending on the desired performance characteristics, allowing the system to adapt to varying input complexity without retraining the base model. This finetuning pipeline is illustrated in Figure \ref{fig:lego-code}.

\section{Evaluation} \label{Sec:Eval}

In this section, we evaluate the predefined methods described in Section \ref{Sec:Approach}. 

\subsection{Research Questions} \label{Subsec:RQs}

We aim to address the following research questions:

\begin{enumerate}

    \item \textbf{RQ1: Does Curriculum Learning perform better than LoRA fine-tuning?} 
    In this research question, we investigate whether curriculum learning strategies outperform LoRA fine-tuning.
    
    \item \textbf{RQ2: What additional benefits does multi-adapter–based curriculum fine-tuning provide beyond overall performance improvement?} 
    This research question examines whether multi-adapter–based curriculum fine-tuning offers advantages beyond overall performance gains.

\end{enumerate}

\subsection{Implementation Details} \label{Subsec:implementation}

To address the above research questions, we conducted a series of experiments on a computer equipped with one NVIDIA GeForce RTX 5090 with 32 GiB of memory. All experiments were implemented using the HuggingFace ecosystem\footnote{https://huggingface.co/}, an open-source platform for natural language processing research.

Specifically, we used: (1) the \textit{Transformers} library, which supports inference, training, and fine-tuning of state-of-the-art pretrained models; (2) the \textit{Datasets} library, which provides a unified interface for accessing and processing widely used NLP datasets; and (3) the \textit{Evaluate} library, which offers standardized tools for computing evaluation metrics across machine learning tasks.

 \subsection{Experiments} \label{Subsec:exps}

 To answer our research questions, we conducted the following experiments:

\textbf{\texttt{EX1:}} In this experiment, we fine-tune \textbf{\texttt{Qwen2.5-Coder-1.5B}} \cite{qwencoder}, a recent coding LLM that demonstrates strong performance across multiple programming tasks, particularly text-to-SQL, using the three fine-tuning strategies detailed in Section \ref{Sec:Approach}. First, we employ LoRA fine-tuning, allowing the dataset to be shuffled without considering the order of samples, as in standard fine-tuning. Second, we apply single-stage curriculum fine-tuning, where we disable shuffling and ensure that the model processes the samples in increasing order of complexity (one epoch). Finally, we employ multi-adapter curriculum fine-tuning with multiple stages, dividing SB-CL into four subsets, EASY, MEDIUM, HARD, and EXTRA, and fine-tuning consecutively on each subset. The choice of a small model is intentional, it enables faster experimentation cycles and lower inference latency, while allowing us to focus on evaluating the proposed method independently of large model scale.

\begin{figure}[htbp]
    \centering
    \includegraphics[width=0.49\textwidth]{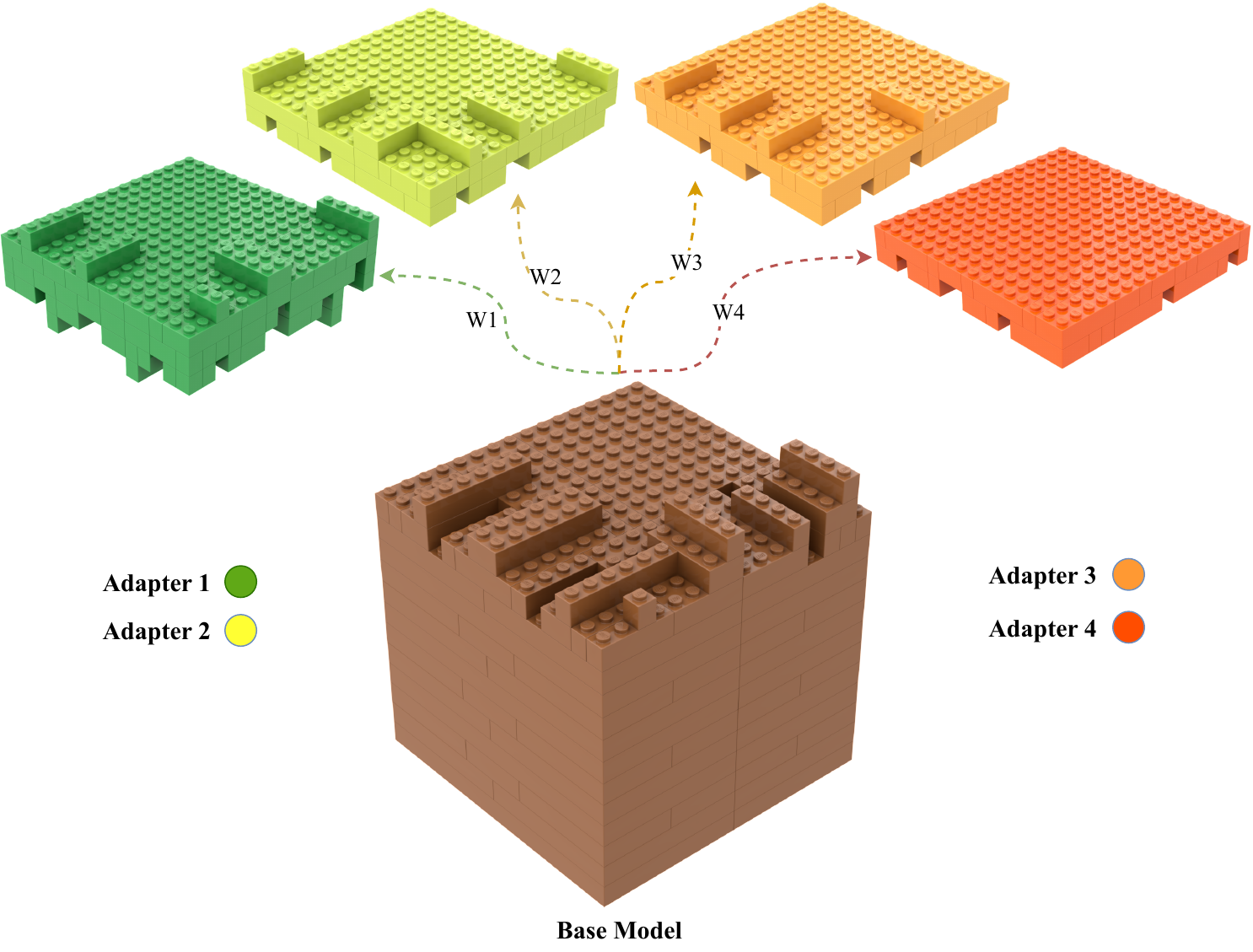}
    \caption{LeGo-Code during Inference}
    \label{fig:lego-code-inference}
\end{figure}

\textbf{\texttt{EX2:}} In this experiment, we evaluate multiple combinations of the adapters obtained from the previous experiment to investigate how LeGo-Code can be applied across different levels of task complexity and to determine which adapter combinations perform best for each level as depicted in Figure \ref{fig:lego-code-inference}. Specifically, we analyze whether composing adapters trained on varying difficulty stages can provide improved performance compared to using a single adapter. This experiment is designed to demonstrate the flexibility and effectiveness of LeGo-Code, highlighting its ability to dynamically assemble specialized components to address tasks of different complexity levels. The results provide insight into how modular adapter composition can enable more efficient SQL query generation at each complexity level.

\subsection{Answers to RQs} \label{Subsec:results}

\subsubsection{\textbf{RQ1}}

\begin{table}[b]
\begin{tabular}{|
>{\columncolor[HTML]{FFFFC7}}l |c|c|}
\hline
\cellcolor[HTML]{FFFE65}\textbf{Model} & \cellcolor[HTML]{FFFE65}\textbf{SPIDER} & \cellcolor[HTML]{FFFE65}\textbf{BIRD} \\ \hline
Qwen2.5-Coder-1.5B & 44.3          & 9.91           \\ \hline
LoRA finetuning  & 57.1          & 16.62          \\ \hline
Single Stage CL    & 34.1          & 13.04          \\ \hline
\textbf{LeGo-Code} & \textbf{59.1} & \textbf{18.90} \\ \hline
\end{tabular}
\caption{Different training strategies Execution Accuracy on SPIDER and BIRD development sets }
\label{tab:execution-accuracy}
\end{table}

Table \ref{tab:execution-accuracy} illustrates the execution accuracy of different training strategies detailed in Section \ref{Sec:Approach}. Qwen2.5-Coder-1.5B-Instruct shows relatively low performance, achieving 44.3\% on the SPIDER development set and 9.91\% on the BIRD development set. LoRA fine-tuning using SB-CL substantially improves performance, yielding gains of +12.8\% on SPIDER and +6.71\% on BIRD. However, unexpectedly, single-stage curriculum learning, where the model is exposed to all SB-CL samples strictly in increasing order of complexity, results in a significant drop in performance on SPIDER (down to 34.1\%), particularly on simple SQL queries, while producing only a modest improvement of +3.13\% on BIRD. This performance remains considerably worse than that of LoRA fine-tuning. LeGo-Code, our approach based on multi-adapter fine-tuning applied sequentially across four SB-CL subsets from easy to extra hard, achieves the best results. It outperforms LoRA fine-tuning by +2.0\% on SPIDER and +2.28\% on BIRD, reaching 59.1\% and 18.90\% execution accuracy, respectively.

Overall, the results demonstrate that staged multi-adapter training with progressively harder curricula is more effective than both standard fine-tuning and single-stage curriculum learning for improving execution accuracy on both benchmarks.

\subsubsection{\textbf{RQ2}}

\begin{figure}[htbp]
    \centering
    \includegraphics[width=0.45\textwidth]{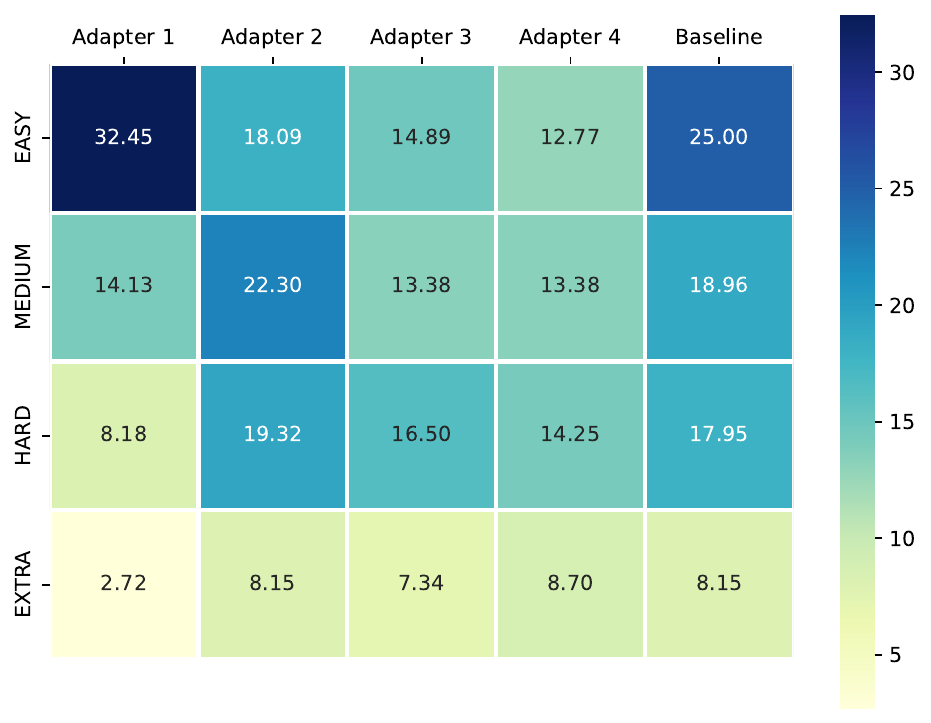}
    \caption{Execution accuracy comparison of various adapters across difficulty levels on BIRD dev.}
    \label{fig:performance_heatmap}
\end{figure}

To investigate whether LeGo-Code provides benefits beyond overall performance improvement, Figure \ref{fig:performance_heatmap} presents the execution accuracy obtained when activating a single adapter at a time for each difficulty level of the BIRD development set. The results reveal strong specialization effects: each adapter performs best on the difficulty level it was trained on, while performance degrades on other levels. For example, Adapter 1, trained on the EASY subset, achieves 32.45\% on EASY queries, substantially outperforming the baseline (25.00\%) by +7.45\%. However, its performance drops sharply on harder subsets (e.g., 2.72\% on EXTRA). Similarly, Adapter 2 performs best on MEDIUM queries (22.30\%), Adapter 3 on HARD queries (16.50\%), and Adapter 4 shows comparatively better results on EXTRA queries (8.70\%). These findings indicate that multi-adapter curriculum fine-tuning enables the model to learn difficulty-specific skills, with each adapter capturing knowledge tailored to a particular complexity level. In contrast, the single baseline model exhibits more uniform but generally lower performance across levels.

Overall LeGo-Code provides modular specialization across difficulty levels. Rather than learning a single compromise solution, the model develops distinct capabilities for easy, medium, hard, and extra-hard queries. This enables more targeted reasoning, reduces interference between tasks of different complexity, and suggests the potential for adaptive inference strategies that select the most appropriate adapter based on query difficulty.

\section{Discussion}\label{Sec:Discussion}

Our evaluation of different training strategies shows that single-stage curriculum learning underperforms LoRA fine-tuning and even degrades performance on SPIDER. This might be due to catastrophic forgetting of simpler SQL patterns as training progresses toward harder examples, as well as insufficient revisiting of earlier samples, an effect mitigated in standard fine-tuning. These results indicate that data ordering alone does not guarantee improved learning. In contrast, LeGo-Code, which applies sequential multi-adapter fine-tuning across difficulty levels, achieves the best overall performance. By allocating separate adapters to different complexity stages, the approach preserves knowledge acquired at each stage while enabling later adapters to focus on more complex reasoning. Adapter-level analysis on the BIRD development set reveals strong specialization: each adapter performs best on queries matching its training difficulty and worse on others, suggesting that the method decomposes the task into difficulty-specific competencies.

\section{Conclusion}\label{Sec:Conclusion}

The text-to-SQL field has seen significant progress in recent years with the introduction of new architectures, datasets, and benchmarks. Despite these advances, state-of-the-art models still struggle with complex, real-world queries over noisy and heterogeneous databases. In this paper, we examined curriculum learning strategies by comparing standard fine-tuning, single-stage curriculum learning, and a proposed multi-adapter approach, LeGo-Code. Our results show that naïve curriculum learning based solely on data ordering can be ineffective and may even degrade performance, likely due to the forgetting of simpler patterns. In contrast, LeGo-Code consistently achieves the best performance across benchmarks by sequentially fine-tuning specialized adapters on progressively more difficult subsets. These findings suggest that curriculum learning becomes substantially more effective when combined with modular adaptation mechanisms rather than relying on ordering strategies alone.

\bibliographystyle{ACM-Reference-Format}
\bibliography{db_references}
\end{document}